\documentclass[sigconf]{acmart}

\AtBeginDocument{%
  }

\copyrightyear{2022}
\acmYear{2022}
\setcopyright{acmlicensed}
\acmConference[PIES-ME '22] {Proceedings of the 1st Workshop on Photorealistic Image and Environment Synthesis for Multimedia Experiments}{October 14, 2022}{Lisboa, Portugal.}
\acmBooktitle{Proceedings of the 1st Workshop on Photorealistic Image and Environment Synthesis for Multimedia Experiments (PIES-ME '22), October 14, 2022, Lisboa, Portugal}
\acmPrice{15.00}
\acmISBN{978-1-4503-9500-7/22/10}
\acmDOI{10.1145/3552482.3556555}
\begin{document}

\title{Language-guided Semantic Style Transfer of 3D Indoor Scenes}


\author{Bu Jin}
\affiliation{%
  \institution{University of Chinese Academy of Sciences}
  \state{Beijing}
  \country{China}}
\email{jinbu18@mails.ucas.edu.cn}

\author{Beiwen Tian}
\affiliation{%
  \institution{Tsinghua University}
  \state{Beijing}
  \country{China}}
\email{tbw18@mails.tsinghua.edu.cn}

\author{Hao Zhao}
\affiliation{%
  \institution{Peking University, Intel Labs}
  \state{Beijing}
  \country{China}}
\email{zhao-hao@pku.edu.cn,hao.zhao@intel.com}

\author{Guyue Zhou}
\affiliation{%
  \institution{Tsinghua University}
  \state{Beijing}
  \country{China}}
\email{zhouguyue@air.tsinghua.edu.cn}

\renewcommand{\shortauthors}{Bu Jin, Beiwen Tian, \& Hao Zhao, Guyue Zhou}
\begin{abstract}
  We address the new problem of language-guided semantic style transfer of 3D indoor scenes. The input is a 3D indoor scene mesh and several phrases that describe the target scene. Firstly, 3D vertex coordinates are mapped to RGB residues by a multi-layer perceptron. Secondly, colored 3D meshes are differentiablly rendered into 2D images, via a viewpoint sampling strategy tailored for indoor scenes. Thirdly, rendered 2D images are compared to phrases, via pre-trained vision-language models. Lastly, errors are back-propagated to the multi-layer perceptron to update vertex colors corresponding to certain semantic categories. We did large-scale qualitative analyses and A/B user tests, with the public ScanNet and SceneNN datasets. We demonstrate: (1) visually pleasing results that are potentially useful for multimedia applications. (2) rendering 3D indoor scenes from viewpoints consistent with human priors is important. (3) incorporating semantics significantly improve style transfer quality. (4) an HSV regularization term leads to results that are more consistent with inputs and generally rated better. Codes and user study toolbox are available at \hyperref[https://github.com/AIR-DISCOVER/LASST]{https://github.com/AIR-DISCOVER/LASST}. 
\end{abstract}




\begin{teaserfigure}
  \centering
  \includegraphics[width=1.0\textwidth]{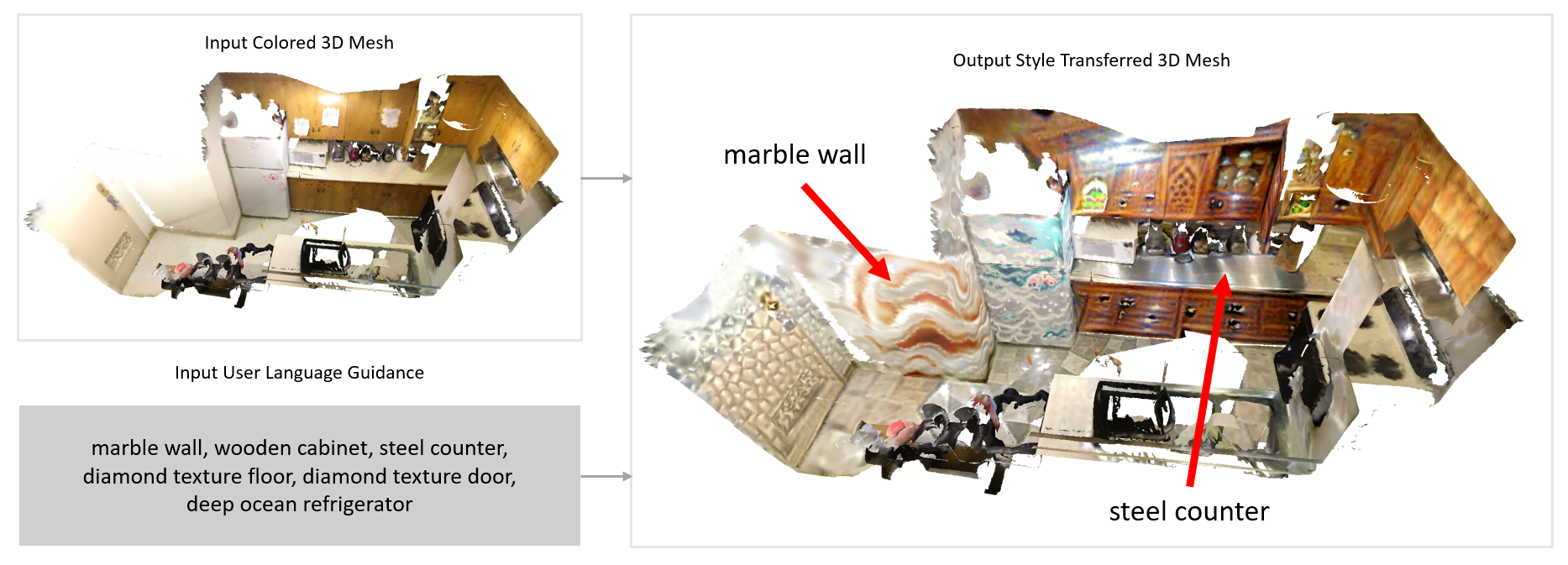}
  \caption{Our algorithm has two inputs: a colored 3D mesh which is the digital twin of a user's room and a set of user-specified text prompts as guidance. The output is a style transferred 3D mesh following the language guidance.}
  \Description{Enjoying the baseball game from the third-base
  seats. Ichiro Suzuki preparing to bat.}
  \label{fig:teaser}
\end{teaserfigure}

\maketitle

\section{Introduction}

3D content creation and editing is a long-existing multimedia demand. With the surge of metaverse, tech giants and consumers are now looking forward to a high-quality virtual world that people can live in and interactive with. We study the problem of 3D indoor scene style transfer, which would promote the user experience of metaverse residents. Shown in Fig.~\ref{fig:teaser}, the system input is a colored 3D mesh as reconstructed by methods like VoxelHashing \cite{niessner2013real} or FlashFusion \cite{han2018flashfusion} and the output is a stylized version corresponding to certain user inputs. We achieve this goal with a new 3D style transfer algorithm with two notable features: language guidance and semantic awareness. Firstly, user inputs are specified by natural language, e.g., \emph{marble wall} and \emph{steel counter} in Fig.~\ref{fig:teaser}. This feature allows open-set arbitrary text inputs that are not restricted to pre-defined style sets. Secondly, our method stylizes different semantic regions separately, allowing a delicate control instead of treating the scene as a whole. 

Language-driven synthesis of 2D images \cite{Patashnik2021StyleCLIPTM} and 3D style transfer via differentiable rendering \cite{Kato2018Neural3M} have both been demonstrated before. Recently, Text2Mesh \cite{Michel2021Text2MeshTN} combines these two methods to build a system that modifies 3D object meshes according to natural language guidance. However, trivially applying the Text2Mesh method to 3D scenes cannot produce satisfactory results due to three reasons. Firstly, the Text2Mesh method randomly samples rendering viewpoints for objects, which is problematic for indoor scenes. The fundamental difference is that objects (e.g., a shoe) can be observed arbitrarily from any viewpoint but an indoor scene should be observed according to human viewing preference priors. Rendering indoor scenes from unnatural viewpoints leads to sub-optimal gradients from the vision-language model (VLM) because the VLM is pre-trained on natural images. Secondly, an indoor scene is composed of many semantic regions and treating the scene as a whole cannot fully leverage the semantic alignment between vision and language. For example, if the \emph{counter} region in Fig.~\ref{fig:teaser} is not known, the \emph{steel} attribute would be transferred to other semantic regions. Thirdly, while transferring the style of an object into an exaggerated one is meaningful in many cases, indoor scene style transfer generally calls for consistent visual experience. A natural idea is to enforce a RGB regularization term between stylized meshes and input meshes. But empirically this loss term cannot effectively guarantee visual consistency.

In order to address aforementioned issues, we propose the LASST algorithm, short for \textbf{la}nguage-guided \textbf{s}emantic \textbf{s}tyle \textbf{t}ransfer of 3D indoor scenes. Firstly, we identify the importance of natural rendering viewpoints and design 6DoF pose priors consistent with human preference, which provide meaningful gradients from the VLM. Secondly, as inspired by 2D semantic style transfer \cite{Lu2017DecoderNO}\cite{lu2018exemplar}, we introduce both ground truth and estimated semantic masks to facilitate aligned feature matching within semantic regions. In the forward pass, the whole scene is rendered, while in the backward pass, we apply the \emph{stop gradient} trick to unrelated semantic regions. Thirdly, we leverage an HSV regularization loss to prevent the scene color from drifting too far from inputs, instead of using a regularization in the RGB space. This alternative leads to stylized meshes that are rated better in A/B user studies.

In order to illustrate the significance of these three points, we benchmark on the ScanNet dataset \cite{dai2017scannet} and SceneNN dataset \cite{hua2016scenenn} with extensive qualitative analyses and A/B user tests. Clear conclusions are reached, showing that users prefer LASST results generated with natural viewpoints, semantic alignment and the HSV regularization. %We also provide other insights on generating visually pleasing results, which are validated by the same A/B user test protocol. 

To summarize, our contributions are as follows:

\begin{itemize}
\item We propose the LASST algorithm. To the best of our knowledge, LASST is the first solution for \textbf{la}nguage-guided \textbf{s}emantic \textbf{s}tyle \textbf{t}ransfer of 3D indoor scenes. It is potentiallly of interest to both multimedia researchers and practitioners. 
\item We demonstrate the importance of using natural rendering viewpoints for indoor scene style transfer.
\item We design a simple but effective strategy based upon the stop gradient trick, to encourage semantic alignment between language guidance and visual data.
\item We propose an HSV regularization loss that leads to results more consistent with inputs.
\item We evaluate our design choices with a large-scale A/B user test and provide an open-source implementation of LASST. 
\end{itemize}

\begin{figure}
  \centering
  \includegraphics[width=0.45\textwidth]{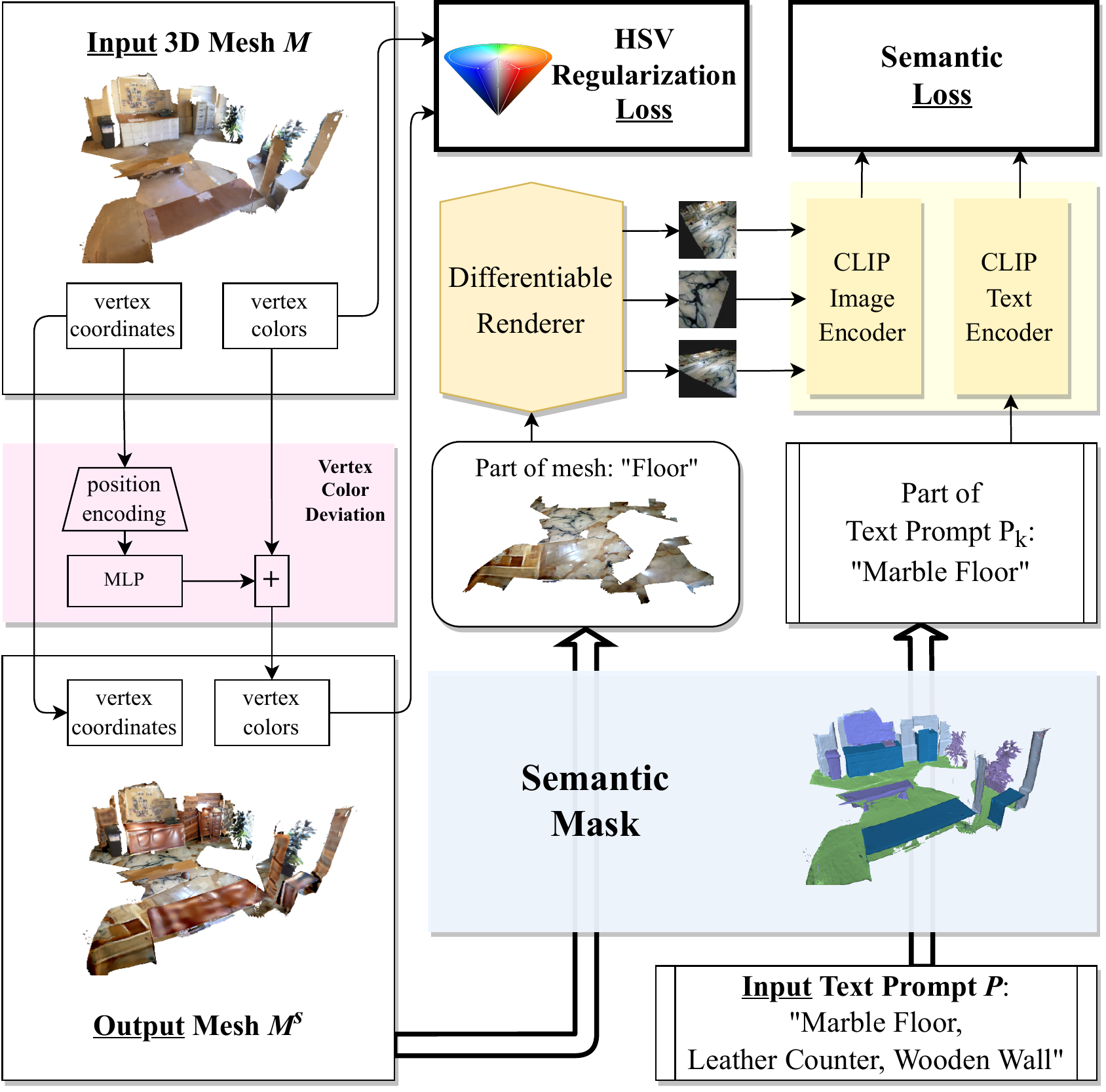}
  \caption{Overview of the LASST algorithm. The input is a colored 3D mesh $\mathbf{M}$ and a text prompt $\mathbf{P}$. The output is a stylized mesh $\mathbf{M^S}$. Input vertex coordinates are sent into a vertex color deviation network to generate color residues, which are added onto the input  vertex color values. Semantic masks are acquired through either accessing the ground truth or running a scene parsing model. The stylized mesh is differentiablly rendered into multiple views and compared with the text prompt of a certain category $\mathbf{P_k}$ through a pre-trained CLIP. The VLM semantic loss and the HSV regularization loss are used to back-propagate gradients to only modify regions corresponding to certain semantics.}
  \Description{Enjoying the baseball game from the third-base
  seats. Ichiro Suzuki preparing to bat.}
  \label{fig:main}
\end{figure}

\section{Related Works}

\textbf{Texturing 3D Meshes.} Successfully texturing 3D meshes requires numerical surface parameterization and state-of-the-art methods pursue conformal mappings \cite{gillespie2021discrete}, global consistency \cite{sharp2018variational} and boundary awareness \cite{li2018optcuts}. However, they do not apply to non-watertight scene meshes as reconstructed using RGB-D streams (e.g. the kitchen in Fig.~\ref{fig:teaser}). Bi \emph{et al.} \cite{bi2017patch} develop a patch search and vote scheme for the texturing of reconstructed objects. Dessein \emph{et al.} \cite{dessein2014seamless} introduce a Poisson regularization into the texture mapping problem on object surfaces. Huang \emph{et al.} \cite{huang20173dlite} trade off geometric details for texture quality using planar primitives. Waechter \emph{et al.} \cite{waechter2014let} design a comprehensive pipeline for large-scale 3D reconstruction texturing, using photometric fidelity and smoothness regularizations. While most methods process readily reconstructed meshes, Zhou \emph{et al.} \cite{zhou2014color} jointly optimize frame poses and texture. Different from these methods, we directly optimize mesh texture using neural network back-propagation, guided by user's natural language input and being aware of semantics.

\textbf{3D Style Transfer.} Kato \emph{et al.} \cite{Kato2018Neural3M} propose a soft surrogate for rasterization so that the rendering of meshes is turned into a differentiable operator. They also demonstrate the application of 3D style transfer with 2D input guidance. Other differentiable renering formulations like OpenDR \cite{loper2014opendr}, Soft Rasterizer \cite{liu2019soft} and DIB-Renderer \cite{chen2019learning} also support 3D style transfer in a similar scheme. 3DStyleNet \cite{yin20213dstylenet} formulates a feed-forward network trained with content and style losses imposed on differentiable renderings. Patch-based methods are also used to stylize 3D models \cite{hauptfleisch2020styleprop}\cite{sykora2019styleblit}. 3D style transfer on point clouds \cite{cao2020psnet} has also been demonstrated. StyleMesh \cite{hollein2021stylemesh} formulates a 3D mesh style transfer algorithm using depth and normal statistics. To the best our knowledge, LASST is the first algorithm that achieve visually pleasing 3D semantic style transfer using language guidance.

\textbf{Language-guided Synthesis.} CLIP \cite{radford2021learning} is a successful vision-language model that models the affinity of RGB images and concepts expressed in texts. Except for its surprising zero-shot classification performance on ImageNet \cite{russakovsky2015imagenet}, CLIP can be used as a \emph{loss function} for visual synthesis. Optimizing the image encoder input to minize its distance with text prompts leads to compelling image generation results \cite{ramesh2021zero}. When combined with diffusion probabilistic models, the visual quality is strengthened \cite{nichol2021glide}. StyleCLIP \cite{Patashnik2021StyleCLIPTM} exploits CLIP text embedding to search for optimization directions in pre-trained StyleGAN models \cite{karras2019style}. Text2Mesh \cite{Michel2021Text2MeshTN} achieves text-driven editing of 3D objects, but has not demonstrate semantic editing results for large indoor scenes. Interestingly, CLIPort \cite{shridhar2022cliport} recently shows that pre-trained VLM can be used to generate manipulation control sequences according to human command.

\section{Method}\label{sec:method}

\subsection{Overview}\label{subsec:Overview}

We give an overview of the LASST algorithm, as illustrated in Fig.~\ref{fig:main}. An indoor scene $\mathbf{M}$ is represented in the 3D mesh format with vertices denoted by $\mathbf{V} \in {\mathbb{R}^{n\times3}}$ and faces denoted by $\mathbf{F} \in {K^{3\times m}},K \in [1,2,...,n]$. %$\mathbf{F} \in {{\{1,...,n\}}^{m\times3}}$.
Each vertex $v_i$ has an initial color and an associated semantic label, which could be ground truth or prediction. The initial colors of a mesh are denoted as $C^{ini}$.
The input text prompt is denoted by $\mathbf{P_{k}}$, where k is the index of sub-prompts and corresponds to a specific semantic category we aim to stylize.
% When we input the text prompt $P$ and target label $l^{in}$, we first compute the semantic mask $Mask\in {{\{0,1\}}^{n\times3}}$, which is later applied to the full mesh, resulting in a masked mesh $M^{Target}$ and remaining mesh $M^{Others}$. 

% We perform style transfer for indoor scenes by optimizing a style transfer network through data-driven optimization. 
We perform style transfer for indoor scenes by optimizing a style transfer network through back-propagation. 
We refer to our style transfer network as VCDN (Vertex Color Deviation Network) as it takes vertex coordinates as input and leverages a multilayer perceptron to predict the color deviation of vertices.
Adding the predicted deviation to the original colors of the input scene $\mathbf{M}$ results in a stylized scene $\mathbf{M}^S$. 
% A Vertex Color Deviation Network (referred to as VCDN) learns the deviation $C^{res}$ of vertex colors from the target style that the text prompt $P$ (e.g., marble floor) promotes, resulting in a stylized scene $M^S$.

Then, a differentiable renderer \cite{KaolinLibrary} is exploited to obtain the rendering results of the stylized scene from several appropriate viewpoints.
To compare the semantics in different modalities, we utilize a pre-trained CLIP model \cite{radford2021learning} to extract the features of the rendered images and the input text prompt.
The negation of cosine similarity between the features are calculated as the loss to optimize (denoted by $\mathscr{L}_{sem}$).
%as it represents the semantic distance between the stylized scene and the target style specified by input text prompt.
% The rendered images and text prompt are  entered to CLIP to calculate the similarity between the stylized mesh $M^S$ and target style.
We also calculate the statistical difference between the stylized and initial scenes in the HSV color space as an additional regularization loss $\mathscr{L}_{hsv}$ to control the deviation from the input scenes. Finally we back-propagate the sum of the losses to optimize VCDN with the CLIP model frozen, while masking out gradients that do not belong to specific semantic regions.
%Ideally the VCDN learns the deviation that produces a stylized scene that could fit the text prompt without overfitting on the bias of CLIP model.

% TODO flag at 4.8 02:23

\subsection{Vertex Color Deviation Network}\label{subsec:VCDN}
As the name suggests, our VCDN predicts the residues $C^{res}$ of vertex colors that deviate from $C^{ini}$. For each scene, we normalize the input vertex coordinates into a unit ball first. Then for each vertex $v_i \in V$, its position encoding $PE(v_i)$ can be computed by:
\begin{equation}
    PE(v_i)=[\cos{(2 {\pi} \textbf{B} v_i)}, \sin{(2 {\pi} \textbf{B} v_i)}]^T
\end{equation}
where $\textbf{B}$ is a random matrix whose values are sampled from Gaussian distributions. The position encoding $PE(v_i)$ is sent to a multi-layer perceptron ${\phi}$ to get the vertex color residues $C^{res}$:
\begin{equation}
    C^{res}=\phi(PE(v_i))
\end{equation}

Finally the stylized mesh $M^S$ is generated by adding the predicted vertex color residues $C^{res}$ to initial vertex colors $C^{ini}$. The vertex color of $\mathbf{M^S}$ is $C^{res}+C^{ini}$. Then the $\mathbf{M^S}$ is passed to the renderer as input to calculate the semantic loss. % and spread it to the faces of the mesh:
%\begin{equation}
%    M^S[C,F_A]=[C^{res}+C^{ini}, %face_{attributes}+Spread(C^{res})]
%\end{equation}

%where $C\in{{(0,1)}^{n\times3\times3}}$ and $F_A \in{{(0,1)}^{n\times3\times3\times3}}$. The $face_{attributes}$ are just composed of (faces number, vertex indices, vertex colors), and the function Spread is to add the $C^{res}$ correspondingly to vertex colors in $face_{attributes}$. 

\begin{figure}
  \centering
  \includegraphics[width=0.4\textwidth]{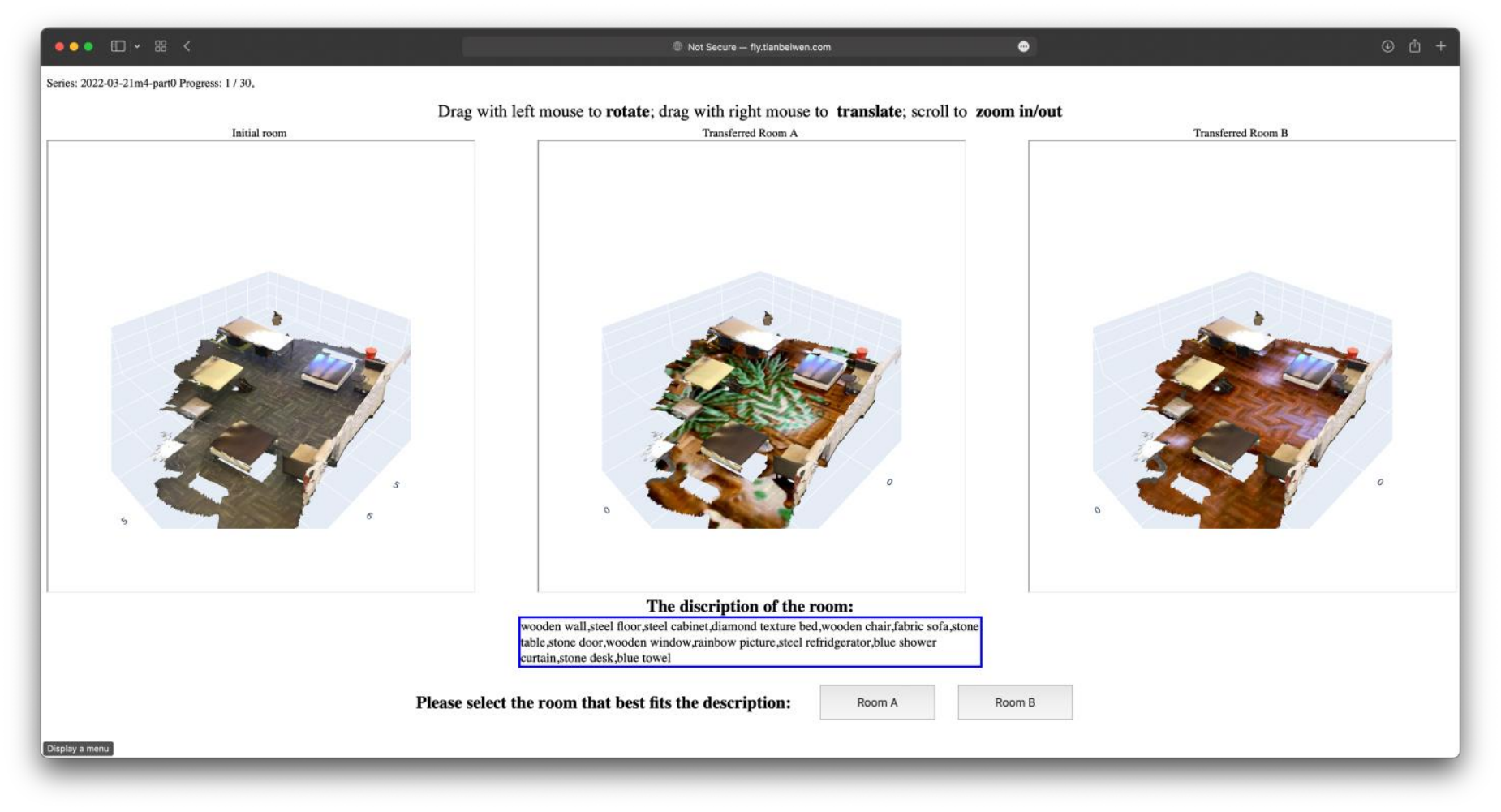}
  \caption{Our online user study interface. Users interact with 3D models and select a result that betters fits the text prompt.}
  \label{fig:userstudy}
\end{figure}

\begin{figure}
  \centering
  \includegraphics[width=0.4\textwidth]{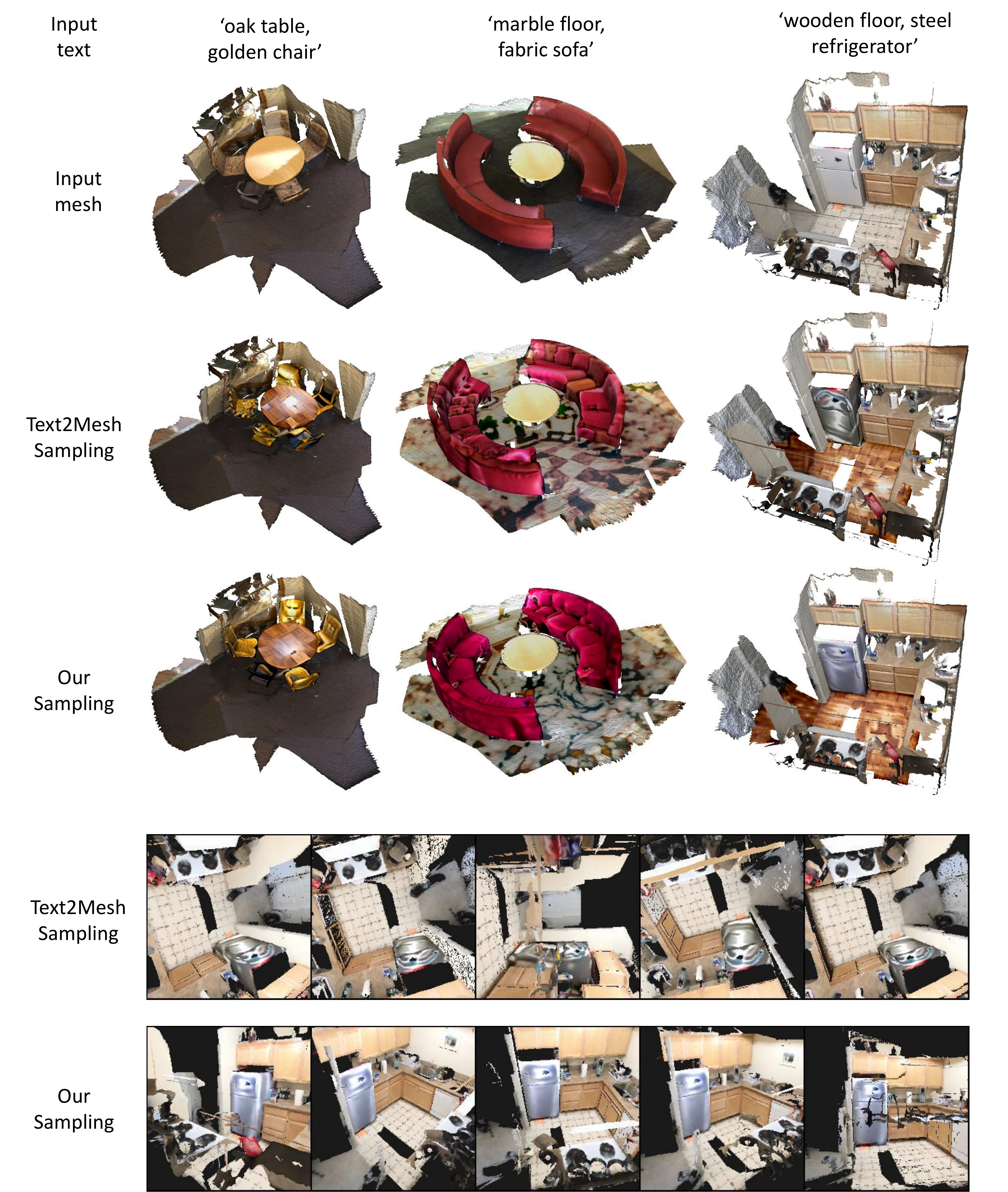}
  \caption{Our sampling strategy generates better results than Text2Mesh's random sampling strategy. The lower two rows are the intermediate rendering results of the scene in the rightmost column, for the \emph{steel refrigerator} prompt.}
  \label{fig:angle}
\end{figure}

\subsection{Semantic Loss via Pre-trained VLM}\label{subsec:renderer}
We define a differentiable renderer $\mathscr{R}$ that renders the 3D stylized mesh $M^S$ into 2D images. To focus on the objects of interest, we render the mesh from views $\theta_j,j=1,...,n_v$ (see Sec.~\ref{subsec:viewpoints}):
\begin{equation}
    I_j=\mathscr{R}(M^S,\theta_j,f_j),j=1,...,n_v
\end{equation}
where $f_j$ is the focal length of the camera, resulting in $n_v$ rendered images $I_j$. To map the rendered images and text prompt $P_k$ to the same high-dimensional space, we exploit a large pre-trained vision-language  model CLIP as our embedding tool. Along with text prompt $P_k$, the rendered images $I_j$ are passed as input to the CLIP model:
\begin{align}
    &S^I=\frac{1}{n_v}\sum_{j}\ E_{img}(I_j)\in {\mathbb{R}^{512}}
    \\
    &S^{P_k}=E_{text}(P_k)\in {\mathbb{R}^{512}}
\end{align}
where $S^I$ and $S^{P_k}$ are in the same multi-modal embedding space the CLIP creates. Then we define the image-text semantic loss $\mathscr{L}_{sem}$ as the opposite number of the cosine similarity between $S^I$ and $S^{P_k}$:
\begin{equation}
   \mathscr{L}_{sem}=-sim_{cos}(S^I, S^{P_k})
\end{equation}

\subsection{Local Awareness}\label{subsec:stopgrad}
% After getting the stylized room and the semantic mask of the input labels $l$, we have two choices to introduce semantic awareness: (i) only rendering a masked mesh; (ii) rendering the whole mesh of the room. Note that in both methods gradient descent can be performed only in the mesh part with label $l$.

% In (i), we only preserve vertices with label $l$ and create a new mesh $M^{l}$ in the same category. Then we render and stylize these objects without its context (other regions in the indoor scene). This strategy is in nature similar to Text2Mesh \cite{Michel2021Text2MeshTN}.

% In (ii), two separate parts of the stylized mesh $M^S$ are both used: mesh with target label $M^l$ and mesh with other label $M^\overline{l}$. 
With the semantic mask of the input labels $l$, the stylized room is separated into two parts: mesh with target label $M^l$
and mesh with other label $M^{\overline{l}}$.
In the forward process, the whole room mesh is passed to the renderer $\mathscr{R}$ to get the images of the room (instead of only some objects in it), see Fig.~\ref{fig:angle}. While in the backward process, the gradient of 
the $M^{\overline{l}}$ is masked out. The whole process can be formulated as:
\begin{equation}
    I_j={\mathscr{R}}(M^l,\theta_j,f_j)+{\mathscr{R}}^{SG}(M^{\overline{l}},\theta_j,f_j) \\
\end{equation}
where $\mathscr{R}$ and ${\mathscr{R}}^{SG}$ are differentiable renderers. The superscript $SG$ denotes Stop Gradient trick.

\subsection{HSV Regularization}\label{subsec:hsv}
The HSV regularization is exploited to prevent the scene color from drifting too far from inputs. For each vertex $v_i$ with RGB color $(R,G,B)$, its color in HSV color space is given by:
\begin{align}
    H&=\left\{
        \begin{array}{rcl}
        0^{\circ},                                     &      & {max=min}\\
        60^{\circ}+\frac{G-B}{max-min}+0^{\circ},      &      & {max=R ,B \leq G}\\
        60^{\circ}+\frac{G-B}{max-min}+360^{\circ},    &      & {max=R ,B > G}\\
        60^{\circ}+\frac{B-R}{max-min}+120^{\circ},    &      & {max=G}\\
        60^{\circ}+\frac{R-G}{max-min}+240^{\circ},    &      & {max=B}
        \end{array}
    \right.
    \\
    S&=\left\{
        \begin{array}{rcl}
        0,          &      & {max=0}\\
        1-\frac{min}{max},  &      & {otherwise}
        \end{array}
    \right.
    \\
    V&=max
\end{align}
where max is the maximum of R,G,B and min is the minimum. Note that $H=0^{\circ}$ and $H=360^{\circ}$ represent the same hue. Given an original mesh and its stylized mesh, we acquire HSV color $(h_1,s_1,v_1)$ and $(h_2,s_2,v_2)$. Then the HSV loss is calculated by:
\begin{equation}
	\begin{split}
    \mathscr{L}_{HSV}=&w_1\mu[(\cos h_1-\cos h_2)+(\sin h_1-\sin h_2)]\\
    &+w_2\mu(s_1-s_2)+w_3\mu(v_1-v_2)
	\end{split}
\end{equation}
where $\mu(x)$ is the average value of $x$ on all vertices and $w_i$ is hyperparameter which controls the weight of HSV loss.

\subsection{Viewpoint Selection}\label{subsec:viewpoints}
As mentioned before, during each optimization procedure we stylize one semantic region in an indoor scene. Therefore we, or the camera, should focus on the target region to prevent the situation that the rendered images fully or largely ignore our targets.

Given a mesh that has been normalized into a unit ball, we first choose a viewpoint $\theta$ and a focal length $f$ randomly. Then we apply random disturbance to the viewpoint and focal length, resulting in $\theta'$ and $f'$. We get the rendered image from renderer defined previously using parameters $\theta'$ and $f'$. Then we compute the ratio $R$ of the target region in the full image and choose $n_v$ images where $R\in (R_{min}, R_{max})$. The $R_{min}$ ensures that an enough portion of pixels of the rendered images display the target region while the $R_{max}$ keep the focal length in a reasonable scale so that most part of the target region is visible. Then we apply $\theta_j$, $f_j$ to the renderer when we train the Vertex Color Deviation Network. 

Additionally, for a camera located in coordinates(x,y,z), the camera pitch angle $\vec{\gamma}$ is computed as:
\begin{equation}
    \vec{\gamma}=(-z\times x, -z\times y, x\times x+y\times y)
\end{equation}
which prevents the room to be rendered from an unnatural angle (e.g., the upside down cases in Fig.~\ref{fig:angle}).

\begin{figure}
  \centering
  \includegraphics[width=0.45\textwidth]{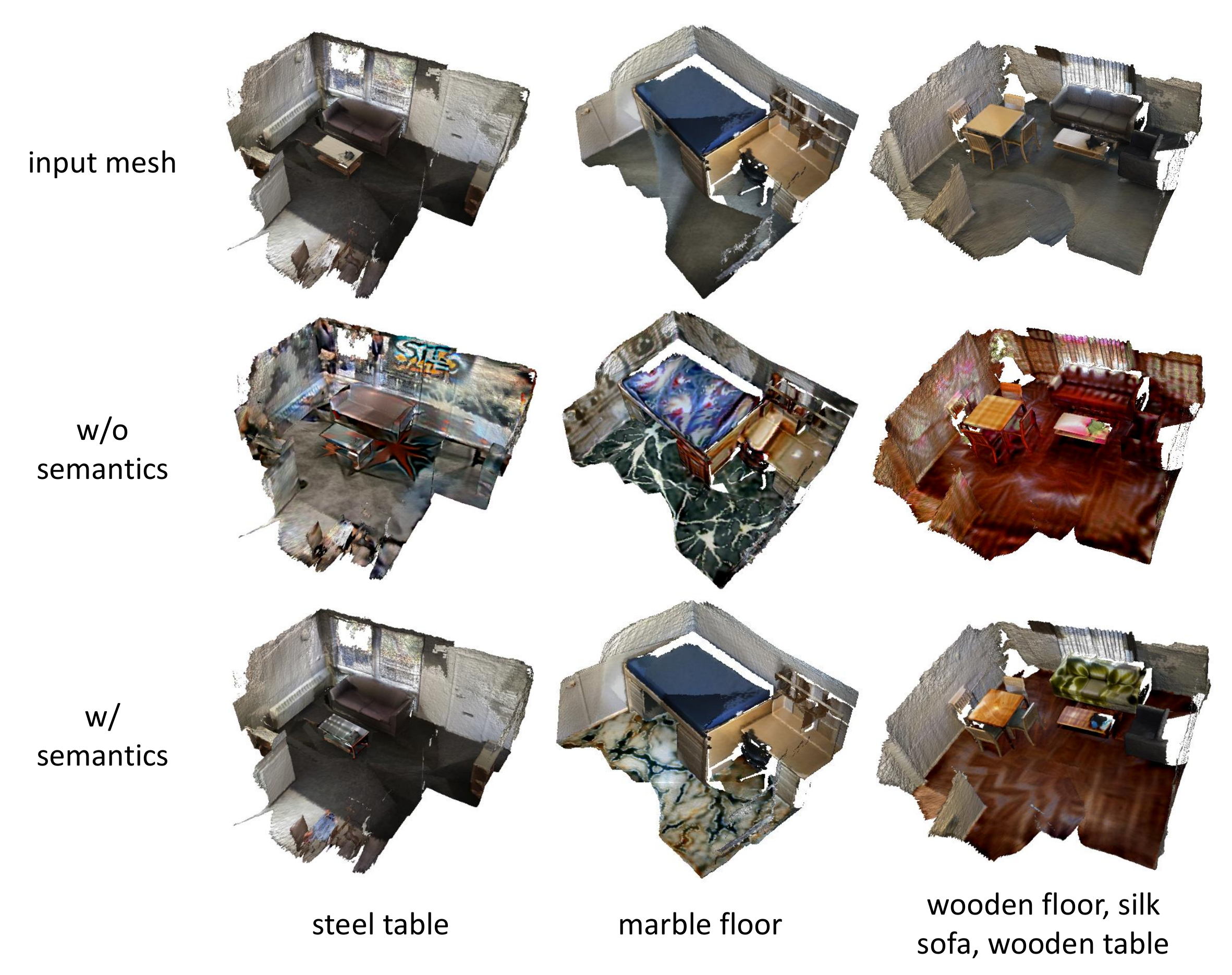}
  \caption{Without semantic region masks, the style prompts are wrongly transferred.}
  \label{fig:mask}
\end{figure}

\begin{figure}
  \centering
  \includegraphics[width=0.35\textwidth]{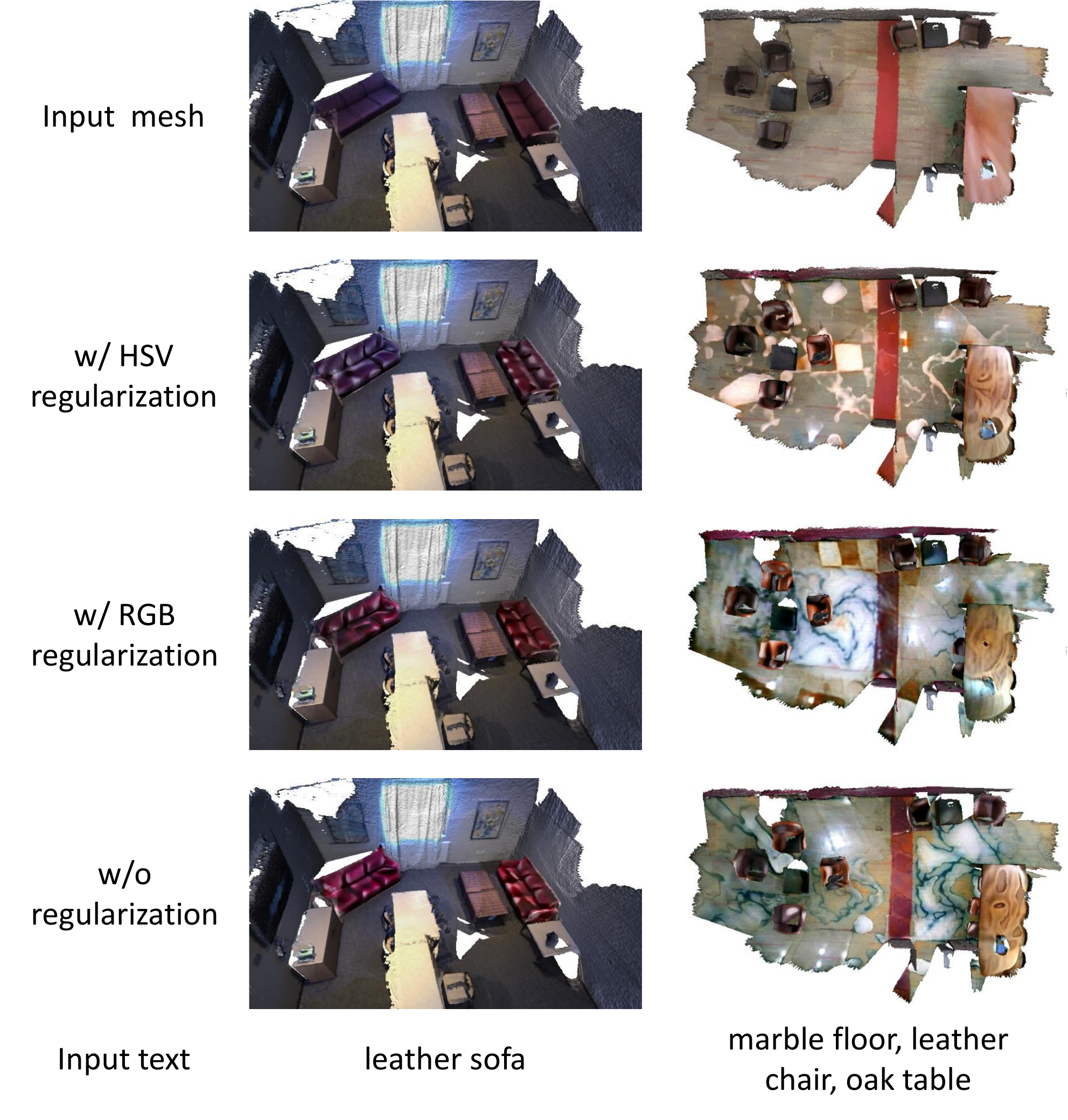}
  \caption{With the HSV regularization term, LASST generates style transfer results that look more similar with the input.}
  \label{fig:hsv}
\end{figure}

\section{Experiments}

\begin{figure*}
  \centering
  \includegraphics[width=0.9\textwidth]{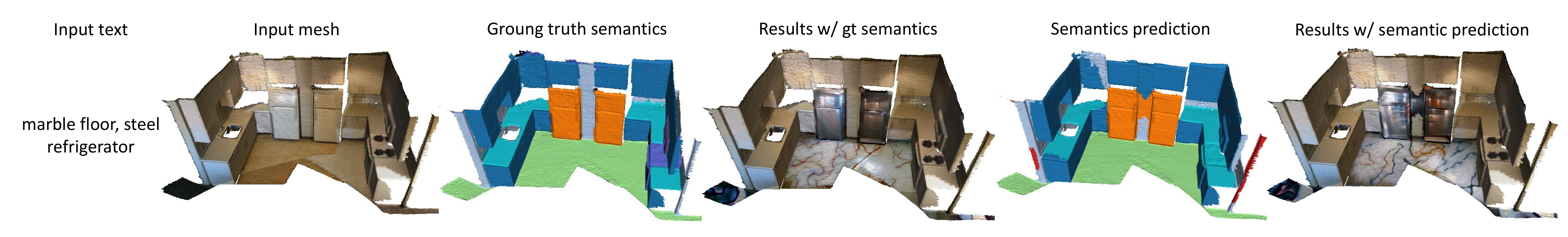}
  \caption{The comparision between stylization with ground truth masks and predicted masks.}
  \Description{Enjoying the baseball game from the third-base
  seats. Ichiro Suzuki preparing to bat.}
  \label{fig:pred}
\end{figure*}

In this section, we introduce the implementation details and experimental designs. Then we provide ablation studies of the proposed design choices, both qualitatively and quantitatively.

\subsection{Dataset}\label{sec:dataset}
The datasets used in our methods include the ScanNetV2 dataset \cite{dai2017scannet} and SceneNN dataset \cite{hua2016scenenn}.
ScanNetV2 is a large-scale RGB-D video dataset with 3D reconstructions of indoor scenes with over 1500 scans.
SceneNN is an RGB-D scene dataset consisting of more than 100 indoor scenes. All scenes are reconstructed into triangle meshes and have per-vertex and per-pixel annotation. 

\subsection{Implementation Details}\label{sec:implementation}

For each semantic part of the scene, we update the VCD network for 700 iterations using an Adam optimizer.

The differentiable renderer used in our framework is provided by Kaolin python library \cite{KaolinLibrary}. For each forward pass, we render the scenes with $n_v=5$ selected viewpoints without shading. The $R_{min}$ and $R_{max}$ are 0.25 and 0.70 respectively in our selection.

As augmentations, we adopt random resizing and cropping to the rendered results and apply random perspective transforms to the images.
Augmented images, along with text prompts, are sent to a pre-trained ViT-B/32 CLIP model to extract features for semantic loss calculation.

The hsv loss weight $(w_1,w_2,w_3)$in our experiments is $(0.2,0.3,0.3)$, which is obtained from user study results. 

\subsection{User Study Protocol} \label{sec:userstudyprotocol}
We conduct the quantitative experiments by user study with an interactive web-based program, which we build to collect human preferences of scene pairs in the ablation studies.
Each scene pair for comparison consists of two scenes stylized from the same initial scene with the same text prompt but with different experiment settings.
Users are asked to selected one that better fits the text prompts from the two stylized scenes. 
During experiments, we divide the scene pairs into groups of 30 to avoid inaccuracy caused by boredom of users. Each of 40 users checked $10\times 30$ pairs of comparisons.

We show the layout of our user study tool in Fig.\ref{fig:userstudy}.
The original scene and two different stylized scenes are displayed as wavefront-format 3D models in the web pages.
By dragging and scrolling, users are free to zoom in or out as well as translate and rotate each one of the models to observe from a better viewpoint.
% The three scenes are referred to as \textit{Initial Scene}, \textit{Transferred Room A}, and \textit{Transferred Room B} respectively and contain no information about experiment settings, avoiding predetermined opinions from users.
% The shared text prompts used for style transfer are shown below the 3D objects.
% Users are able to click on one of the two buttons marked as \textit{Room A} and \textit{Room B} to show their preferences for the two rooms.
% After clicking, the page is redirected to the next page containing another pair until the end of the group.

\subsection{Ablation Study}\label{sec:ablationstudy}
% In this section, we show ablation studies on the key elements of our LASST algorithm: viewpoint selection with prior, semantic style transfer and HSV regularization loss. We also demonstrate the cases in which estimated scene parsing results are used.
We conduct comprehensive ablation study on multiple datasets to investigate the capability of our LASST algorithm.

\textbf{Viewpoint Selection.}
We investigate the effects of our viewpoint sampling strategy compared with the strategy in Text2Mesh \cite{Michel2021Text2MeshTN} with random camera position and fixed camera pitch direction, as is shown in Fig.~\ref{fig:angle}. In the lower two rows, we visualize intermediate rendering results. Our renderer focuses on the target object with input label (e.g., refrigerator), while the Text2Mesh sampling ignores the position and pose of the target, leading to some rendered images under unnatural viewpoints, which deviate far from natural images used for CLIP pre-training. Also in Fig.~\ref{fig:angle} some results are displayed and we can see clearly our sampling method improves the final stylized room. As shown in Table.~\ref{tab:user}, 78.95\% of user ratings prefer LASST sampling over Text2Mesh sampling. 

\textbf{Semantic Mask.}
In our method, we exploit semantic masks to apply stylization in specific regions of the room, instead of the whole room. The semantic mask helps LASST better align vision and language and stylize the room more precisely. 
For example in the first column of Fig.~\ref{fig:mask}, given an initial room mesh and text prompt \emph{steel table}, we expect a room with a steel table like the third row. But in the second row the whole room is stylized, which fails to meet the user's demand. Moreover, graffiti-like english characters \emph{STEEL} are attached to the wall. This is not surprising as vision language models can understand characters in images, as demonstrated in the \emph{OPENAI written on it} demo in https://openai.com/blog/dall-e/. Also, we try to style transfer the room with several prompts like "wooden floor, silk sofa, wooden table", which is shown in the last column of Fig.~\ref{fig:mask}. Then whole room is stylized as a wooden room even if we have "silk sofa" in text prompt, which shows the stylization would miss some information in the text prompts without semantic mask. As shown in Table.~\ref{tab:user}, 86.98\% of user ratings prefer results generated with semantic masks than without.

\begin{table}[]
\caption{Quantitative A/B user study results. The number means percentage of user tests that prefer A over B.}
\begin{tabular}{ccc}
\toprule
A/B Preference Rate                                                              & ScanNet  & SceneNN             \\ \midrule
Sampling(LASST v.s. Text2Mesh)    & 78.95\%  & 76.04\%             \\ \midrule
Semantic Mask(w/ v.s. w/o)        & 86.98\%  & 85.45\%             \\ \midrule
Semantic Mask(GT v.s. Prediction) & 71.43\%  & 59.09\%             \\ \midrule
Regularization(HSV v.s. None)     & 64.80\%  & 65.07\%             \\ \midrule
Regularization(HSV v.s. RGB)      & 60.51\%  & 60.27\%             \\ \bottomrule
\end{tabular}

\label{tab:user}
\end{table}

\begin{table}[]
\caption{CLIP Score and Runtime Comparison. CLIP scores measures the compatibility between an image and a caption by cosine similarity.}
\begin{tabular}{cccc}
\toprule
Regularization  & Sampling      & CLIP Score        & Optimization Time/s       \\ \midrule
None            & LASST         & 0.2972            & 375.53                    \\ \midrule
None            & Text2Mesh     & 0.2744            & 355.27                    \\ \midrule
RGB             & LASST         & 0.2887            & 395.52                    \\ \midrule
HSV             & LASST         & 0.2669            & 385.84                    \\ \bottomrule

\end{tabular}

\label{tab:clipscore_runtime}
\end{table}

\textbf{HSV Regularization.}
Our method uses HSV regularization to encourage consistency with initial scenes. We compare settings applying style transfer with or without HSV regularization.
As is shown in Fig.~\ref{fig:hsv}, we find that the stylized room with HSV regularization well maintains the color impression of input scenes and still achieves good similarity with text prompts. For example given a room with text prompt \emph{leather sofa} (see the first column of Fig.~\ref{fig:hsv}), both methods can stylize the input sofa as leather sofa. However, the result without regularization makes the sofa too bright and stand out in the scene context. As shown in Table.~\ref{tab:user}, 64.85\% of user ratings prefer results generated with HSV regularization than without.

Additionally, we also compare the style transfer effects with RGB regularization. As shown in Table.~\ref{tab:user}, 60.51\% of user ratings prefer results generated with HSV than RGB regularization.

\textbf{Using Estimated Semantic Masks.}
In previous experiments, we conduct our experiments with the ground truth semantic masks. However, when stylizing an indoor scene in real multimedia applications, we need to first conduct semantic segmentation for the room. Thus we investigate the differences between the results with ground truth labels and estimated labels.

The room mesh is first passed as input to a semantic segmantation model, e.g. MinkowskiNet\cite{choy20194d}\cite{luo2021pointly}. Then we exploit the output predicted labels and ground truth labels respectively in our framework. The semantic segmantation results and final stylized results are displayed in Fig.~\ref{fig:pred}. We can see that the results with ground truth semantics and results with semantic prediction are generally the same, despite some regions with wrongly predicted labels (e.g., the wall between two refrigerators).
% The second row shows that when the accuracy of semantic segmentation is not satisfactory, style transfer with correct labels still gives good results.
As shown in Table.~\ref{tab:user}, 71.43\% of user ratings prefer results generated with ground truth labels than estimated labels.

\subsection{CLIP Score and Runtime Comparison}

Although A/B user test is the golden evaluation standard for multimedia applications, we also report the CLIP score after convergence in several settings. As shown in Table.~\ref{tab:clipscore_runtime}, using our sampling strategy can improve the CLIP score from 0.2744 to 0.2972, when compared with Text2Mesh random sampling. This is understandable, because randomly sampled viewpoints leads to unnatural views that is \emph{unfamiliar} to the CLIP model. Meanwhile, using a regularization (RGB or HSV) term inevitably impacts the main loss, but according to our user study the user preference is improved.

On average, the stylization of a room mesh in the ScanNet dataset or SceneNN dataset takes about six minites on a single Nvidia V100 GPU. The optimization time is listed in Table.~\ref{tab:clipscore_runtime}. When compared the optimization time in different settings and find that generally our design choices improve the style transfer quality with limited impact on algorithm speed. %We observe that our strategies have almost the same training time as those without strategies while slightly increasing preprocess time (compared with training time). This demonstrates that we can improve the stylized results with little increase time, making it feasible and extensible to implement these strategies.

\section{Conclusion}

In this paper we propose the first language-driven semantic style transfer algorithm for 3D indoor scenes, named LASST. The inputs are a 3D indoor scene mesh and several phrases specifying target styles. The mesh is rendered differentiablly into multiple 2D images and compared with text prompts in the feature space of a pre-trained vision language model. We identify the importance of better vision language alignment through semantic masks, a viewpoint sampling strategy that incorporates human viewing priors, and an HSV regularization loss that discourages too much drift from input colors. With a large scale A/B user tests, we demonstrate that our design choices are well recognized. We also provide comprehensive qualitative results showing the pros and cons of our method.

\bibliographystyle{ACM-Reference-Format}
\balance
\bibliography{ref}

\end{document}